\documentclass[preprint]{elsarticle}
\usepackage{geometry}
\usepackage{graphicx}
\usepackage{float}
\usepackage{subcaption}
\usepackage{array}
\begin{document}
\pagestyle{empty}

{\Large
\textbf\newline{Automated Radiological Report Generation For Chest X-Rays With Weakly-Supervised End-to-End Deep Learning}
}

Shuai Zhang$^{1,\dagger}$,
Xiaoyan Xin$^{2,\dagger}$,
Yang Wang$^2$,
Yachong Guo$^{1,3}$,
Qiuqiao Hao$^4$,
Xianfeng Yang$^2$,
Jun Wang$^{1,3,5}$,
Jian Zhang$^{1,3,5,*}$,
Bing Zhang$^{2,*}$,
Wei Wang$^{1,3,*}$

$^1$ School of Physics, Collaborative Innovation Center of Advanced Microstructures, Nanjing University, Nanjing, China

$^2$ Department of Radiology, Drum Tower Hospital,  Nanjing University, Nanjing, China

$^3$ Institute for Brain Sciences, Nanjing University, Nanjing, China

$^4$ Institute of Dermatology, Chinese Academy of Medical Sciences and Peking Union Medical College, Nanjing, China

$^5$ State Key Laboratory for Novel Software Technology, Nanjing University, Nanjing, China

\bigskip

$^\dagger $ S. Zhang and X. Xin contribute equally.

$^*$ Corresponding authors: jzhang@nju.edu.cn, zhangbing\_nanjing@nju.edu.cn, wangwei@nju.edu.cn.


\newpage
\section*{Abstract}

The chest X-Ray (CXR) is the one of the most common clinical exam used to diagnose thoracic diseases and abnormalities. The volume of CXR scans generated daily in hospitals is huge. Therefore, an automated diagnosis system able to save the effort of doctors is of great value.
At present, the applications of artificial intelligence in CXR diagnosis usually use pattern recognition to classify the scans. However, such methods rely on labeled databases, which are costly and usually have large error rates.
In this work, we built a database containing more than 12,000 CXR scans and radiological reports, and developed a model based on deep convolutional neural network and recurrent network with attention mechanism. The model learns features from the CXR scans and the associated raw radiological reports directly; no additional labeling of the scans are needed. The model provides automated recognition of given scans and generation of reports.
The quality of the generated reports was evaluated with both the CIDEr scores and by radiologists as well. The CIDEr scores are found to be around 5.8 on average for the testing dataset. Further blind evaluation suggested a comparable performance against human radiologist.

\section*{Introduction}

The chest X-Ray (CXR) is one of the most common diagnostic techniques for respiratory system. It iss quick and inexpensive, and yields low radiation. The volume of daily CXR scans in hospitals is huge and their examination and interpretation consume lots of time and effort of radiologists. Therefore, it is desirable to develop an automated system that is able to examine and interpret CXR radiographs automatically. Moreover, an automated system may help reduce inter-observer variations due to the factors including individual experience, quality of the radiograph, time and personality type \cite{tudor1997assessment}. The adoption of an automated system will \b{lead} to a more standardized terminology and treatment, and benefit the collaborations between different parties. The system may further evoke new applications such as remote diagnosis, self-service diagnosis and so on.

Previously, many works focused on automated classification of the CXR scans. These works are usually based on variants of Convolutional Neural Networks (CNN) and supervised learning \cite{Rajpurkar2017CheXNet,Ypsilantis2017Learning,Pesce2017Learning,islam2017abnormality,Yao2017Learning,Yan2018Weakly,Guan2018Diagnose,Rubin2018Large}. However, at least two problems hinder the practical applications of automated systems in hospital. First, the sensitivity and false positive rates of these classification approaches seem to be saturated. Further improvement may need significant increase the amount and quantity of labeled samples, which are very expensive in the medical field. Second, the decision strategy underlying these systems has not been well understood, making it difficult to track the errors and gain the trust of doctors and patients.

In this work, we developed a model based on deep recurrent neural networks with attention mechanism that learns from CXR images and the raw radiological reports simultaneously. The deep neural networks have shown great potential in characterizing and classifying complex data in a broad range of fields \cite{goodfellow2016deep,Silver2016Mastering,Hui2015RNA}.
After training with our database, the neural network is able to automatically generate radiological reports for given scans. Our work has the following novel features.
First, the training of our model is only weakly supervised; the model directly learns from the image and the raw radiological reports stored in the hospital database; no further classification and labeling of the images by human is required. That is, in contrast to most machine learning models. This feature greatly facilitates the acquisition of data and training of large-scale models. Second, instead of a simple classification of the case into one or several disease categories, the output of our model is a descriptive report regarding different conditions of the chest; the output is directly readable for patients. Third, the implementation of the attention mechanism adds another level of understanding of how the model works, facilitating debugging and optimizing of the model.
In the following sections, we describe the model architecture, training and testing procedures, and the performance evaluated with the CIDEr scores and by human radiologists.

\section*{Methods}

\subsection{The architecture of the network}

\begin{figure}[h]
\centering
\includegraphics[width=1\textwidth]{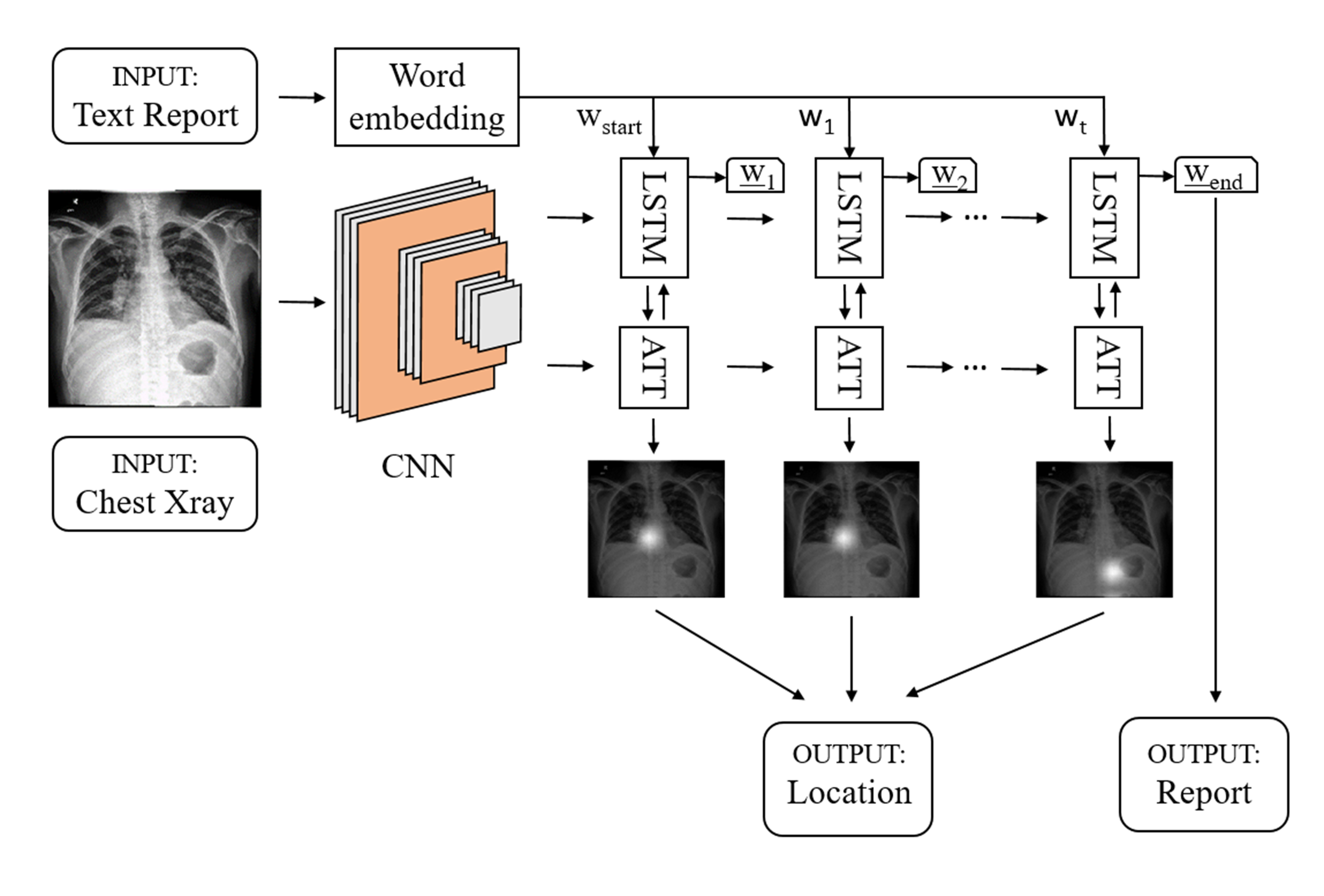}
\caption{The architecture of the whole network. During training, the inputs are the CXR image and the associated text report. The image is encoded by the CNN module and fed to the LSTM with attention mechanism for generating text report.}
\label{fig1}
\end{figure}

Figure \ref{fig1} shows the overall architecture of the neural network. During training, the neural network reads in both CXR images and the raw radiological reports, and outputs human readable texts. The output is then compared with the ground truth to calculate the loss function, which is minimized with the gradient back-propagation technique. After training, the model is able to automatically generate reports for given CXR images.
The design of the model architecture was inspired by the pioneer work of Xu et al. \cite {xu2015show}, where they developed an RNN to generate captions for daily images, such as those in Flickr and MS COCO databases. The model is also similar to those by Zhang et al. \cite{Zhang2017Mdnet} and Wang et al. \cite{Wang2018TieNet} in terms of the purpose of automatically generating medical reports. However, the architecture of our model was redesigned to better fit the organization of our database.

The model contains a 121-layer Densely Connected Convolutional Network (DenseNet) \cite{Huang2017Densely}, which is used as a visual information encoder to extract features from the input images. The encoder is composed of four blocks; each block contains several convolutional layers, each takes all preceding feature-maps as inputs. The blocks are connected by transition layers. According to Huang et al., DenseNets alleviate the vanishing gradient problem, strengthen feature propagation, encourage feature reuse, and substantially reduce the number of parameters \cite{Huang2017Densely}. Compared with many other CNNs, they converge faster and are appropriate for smaller datasets. Therefore, DenseNets are suitable for medical images. The output of the last layer of the DenseNet block is fed to the Long Short-Term Memory (LSTM) network to generate descriptions for the given CXR image.

The LSTM network \cite{Hochreiter1997Long} is adopted to generate texts for the given CXR image word by word. At each step, it reads the output of the last layer of the DenseNet block and the previous generated word, and outputs the next word. In detail, our LSTM implementation follows that of \cite{zaremba2014recurrent}, i.e.,

\begin{equation}
\left( {\begin{array}{*{20}{l}}
{{{\rm{i}}_t}}\\
{{{\rm{f}}_t}}\\
{{{\rm{o}}_t}}\\
{{{\rm{g}}_t}}
\end{array}} \right) = \left( {\begin{array}{*{20}{c}}
\sigma \\
\sigma \\
\sigma \\
{\tanh }
\end{array}} \right){W_{h,e + c + h}}\left( {\begin{array}{*{20}{c}}
{E{y_{t - 1}}}\\
{{V_{gav}}}\\
{{h_{t - 1}}}
\end{array}} \right),
\end{equation}

\begin{equation}
{c_t} = {{\rm{f}}_t} \bigodot {c_{t - 1}} + {i_t} \bigodot {g_t},
\end{equation}

\begin{equation}
{h_t} = {o_t} \bigodot \tanh ({c_t}),
\end{equation}
where $i_t$, $f_t$, $c_t$, $o_t$, $h_t$ are the input, forget, memory, output and hidden gates of the LSTM, respectively. $W$ is the weight matrix, $\sigma$ the logistic sigmoid function, $V_{gav}$ the global average of DenseNet output, $y_{t-1}$ the previous generated word, and $E$ is an embedding matrix. The symbol $\bigodot$ denotes element-wise multiplication. The subscript $h$ is the size of the hidden states, $e$ the vocabulary size, and $c$ is the channel number of the output of DenseNet.

Attention mechanism has been widely adopted in visual image processing since it improves the model performance and adds a level of understanding of how the model works. It mimics the human visual attention mechanism by learning to focus on a certain image region. Specifically, a soft attention mechanism is implemented in the model, which calculates a set of weights conditioning on the image representation and on the hidden state. These weights are multiplied with the output vectors of the DenseNet to get a weighted representation of the image, which is then utilized by the recurrent neural network to generate descriptions. The corresponding equations are
\begin{equation}
{\alpha _t} = {\rm softmax}({W_{p,h}}{h_t} + {W_{1,c}}V),
\end{equation}
\begin{equation}
{C_t} = V{\alpha _t},
\end{equation}
\begin{equation}
{{\bf{y}}_t} \sim {{\bf{p}}_t} \propto {\rm{exp}}\left( {{{\rm{W}}_{e,h}}{{\rm{h}}_t}{\rm{ + }}{{\rm{W}}_{e,c}}{{\rm{C}}_t}} \right),
\end{equation}
where $\alpha _t$ is the attention weight, $V$ the output of DenseNet, $C_t$ the weighted context vector, $\bf{p}_t$ the probability distribution of the words, and $\bf{y}_t$ is the predicted word sampled from $\bf{p}_t$.

The loss function is the cross entropy between the ground truth and the prediction distributions of the texts,
\begin{equation}
loss({\bf{p}},{\bf{y}}) =  - \frac{1}{l}\sum\nolimits_{j = 1}^l {log\left( {{{\bf{p}}_j}({{y}_j})} \right)},
\end{equation}
where $\bf{p_j}$ is the probability distribution predicted at the $j$-th step, ${y_j}$ the index of the $j$-th word in the ground truth text, and $l$ the length of the text.

\subsection{Datasets}

\begin{figure}[!ht]
\centering
\includegraphics[width=1\textwidth]{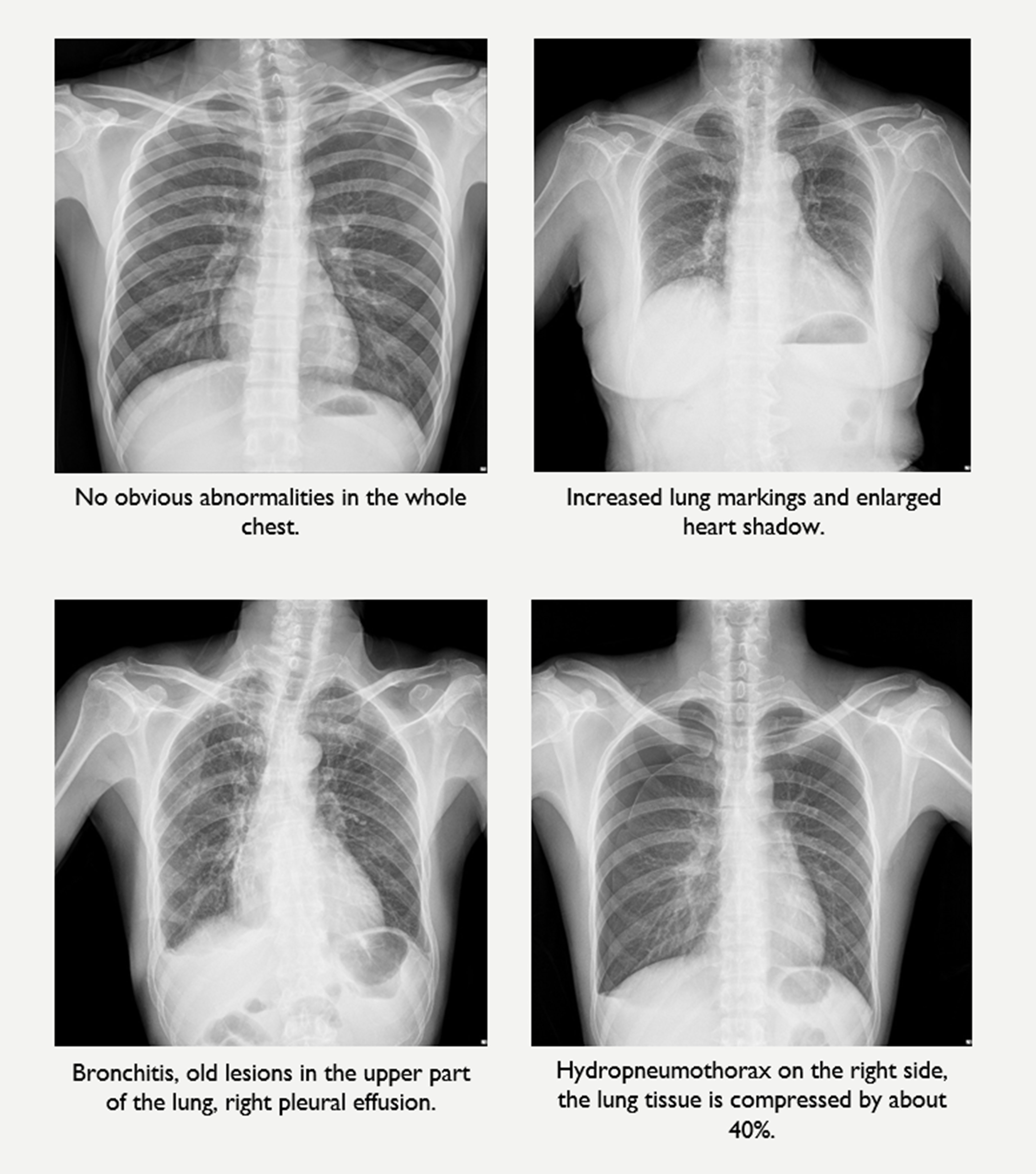}
\caption{Four samples chosen from the dataset. Note the English description is translated from Chinese by human.}
\label{fig2}
\end{figure}

All the chest X-ray scans and the associated radiological reports were provided by Nanjing Drum Tower Hospital. In total the dataset contains 12, 219 images and the same number of reports in Chinese language. All reports were investigated by one expert (of attending doctor level or above) and double checked by another expert (of associate chief physician level or above). The dataset was randomly split into three sets, 80\% samples for training, 10\% for validating, and 10\% for testing. Figure \ref{fig2} shows some examples randomly taken from the dataset.
Since there are no blanks between Chinese words, the python module jieba \cite{jieba} was used for text segmentation. After processing all the radiological reports in the dataset, a vocabulary of size 424 was obtained. The words were represented with one-hot-vectors. The words that appeared less than three times were represented with a special token $<$nou$>$. Two other special tokens were $<$start$>$ and $<$end$>$, indicating the beginning and ending of the reports, respectively.

\subsection{Training Procedures}

Transferring learning technique was employed to speed up the convergence of training. Specifically, the 121-layer DenseNet was pre-trained with the ChestX-ray8 dataset released in September 2017 \cite{Wang2017ChestX} for a classification task in a supervised way. The training procedure was similar to that in \cite{Rajpurkar2017CheXNet}. The ChestX-ray8 dataset contains 110k chest x-ray images and 14 types of diseases labels. The resulted weights were then transferred to the encoder module of the model. In the training process that followed, the parameters in the first two dense blocks were fixed while that in the others were fine-tuned by the gradient back-propagation algorithm.

During training, the original x-ray images were resized to 256*256 pixels and then processed with histogram equalization to increase the contrast. The size of the hidden unit of LSTM was set to 512 and the embedding size was 256. Adam optimizer was used for the optimization process. The learning rate of the DenseNet was set to $1.0 \times 10^{-4}$ and that of the LSTM was set to $5.0 \times 10^{-4}$.

\subsection{Evaluation metrics}

In order to evaluate for image $I_i$ how well a generated sentence $c_i$ matched the consensus of a set of descriptions $S_i=\{s_{i1}, ..., s_{im}\}$, the Consensus-based Image Description Evaluation score (CIDEr) \cite{Vedantam2015CIDEr} was used.

To calculate the CIDEr score, the Term Frequency Inverse Document Frequency (TF-IDF) weighting ${g_k}({s_{ij}})$ for each $n$-gram $\omega_k$ in the sentence $s_{ij}$ was first calculated as
\begin{equation}
{g_k}({s_{ij}}) = \frac{{h_k}({s_{ij}})}{{\sum _{\omega l \in \Omega }}{h_l}({s_{ij}})}\log \left( {\frac{{\left| I \right|}}{\sum\nolimits_{{I_{p \in I}}} {\min (1,\sum\nolimits_q {{h_k}({s_{pq}})} )} }} \right),
\end{equation}
where $h_k({s_{ij}})$ is the number of times an $n$-gram $\omega_k$ occurs in the sentence $s_{ij}$, $\Omega $ the vocabulary of all $n$-grams and $I$ the set of all images in the dataset.

Then the CIDEr$_n$ score for $n$-grams of length $n$ was calculated as
\begin{equation}
{\rm CIDEr}_n(c_i,S_i) = \frac{1}{m}\sum\nolimits_j {\frac{{{g^n}({c_i}) \cdot {g^n}({s_{ij}})}}{{\left\| {{g^n}({c_i})} \right\|\left\| {{g^n}({s_{ij}})} \right\|}}},
\end{equation}
where $g^n(s_{ij})$ is a vector formed by $g_k(s_{ij})$ corresponding to all $n$-grams of length n. $g^n(c_i)$ is similarly defined for the generated sentence $c_i$.

At last, the CIDEr score was calculated as the average over all $n$-grams,
\begin{equation}
{\rm CIDEr}(c_i,S_i) = \frac{1}{N} \sum_{n=1}^N {\rm CIDEr}_n (c_i,S_i).
\end{equation}

\section*{Results}

\begin{figure}[!ht]
	\begin{center}
	\includegraphics[width=0.48\textwidth]{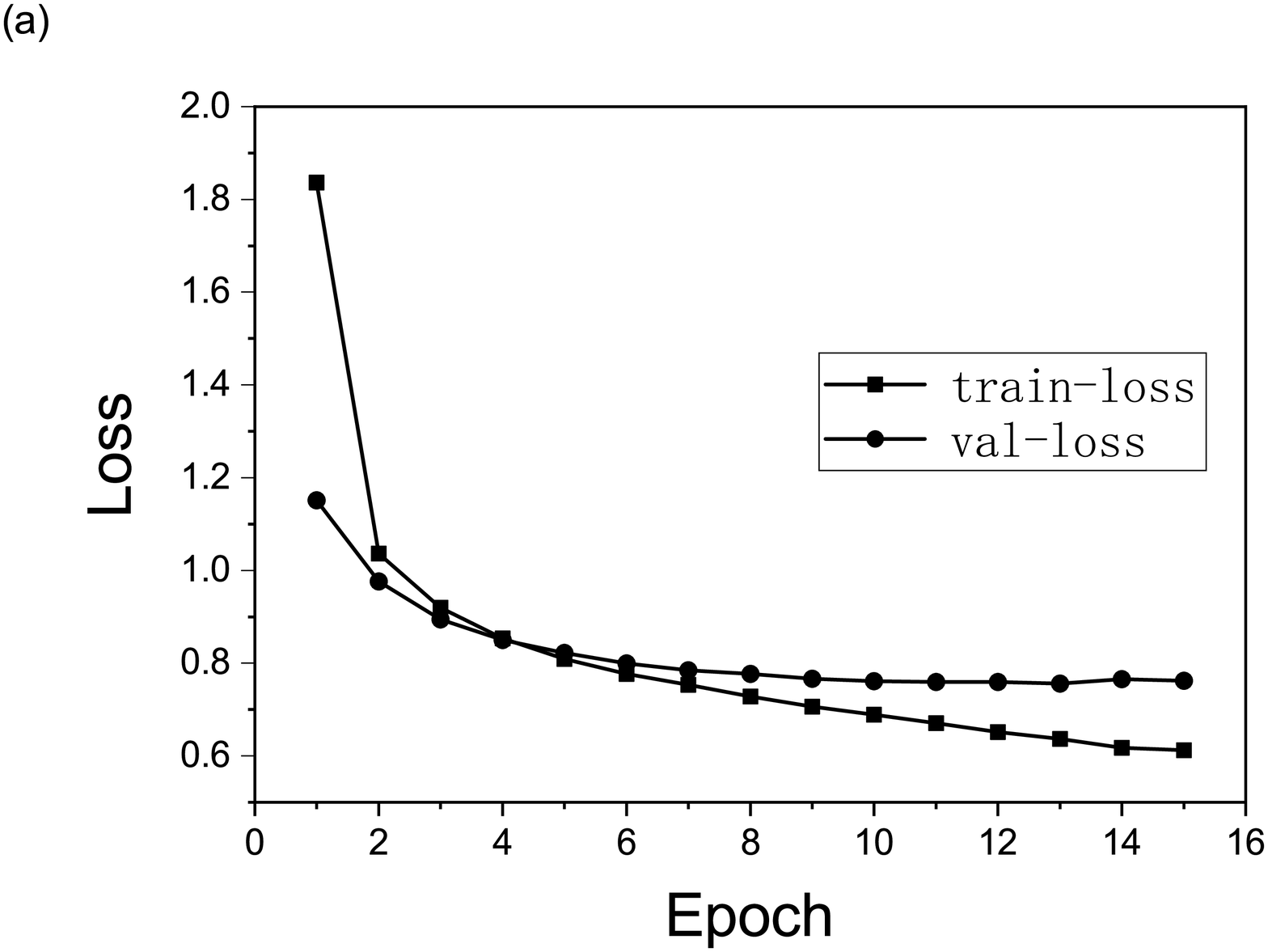}
	\includegraphics[width=0.48\textwidth]{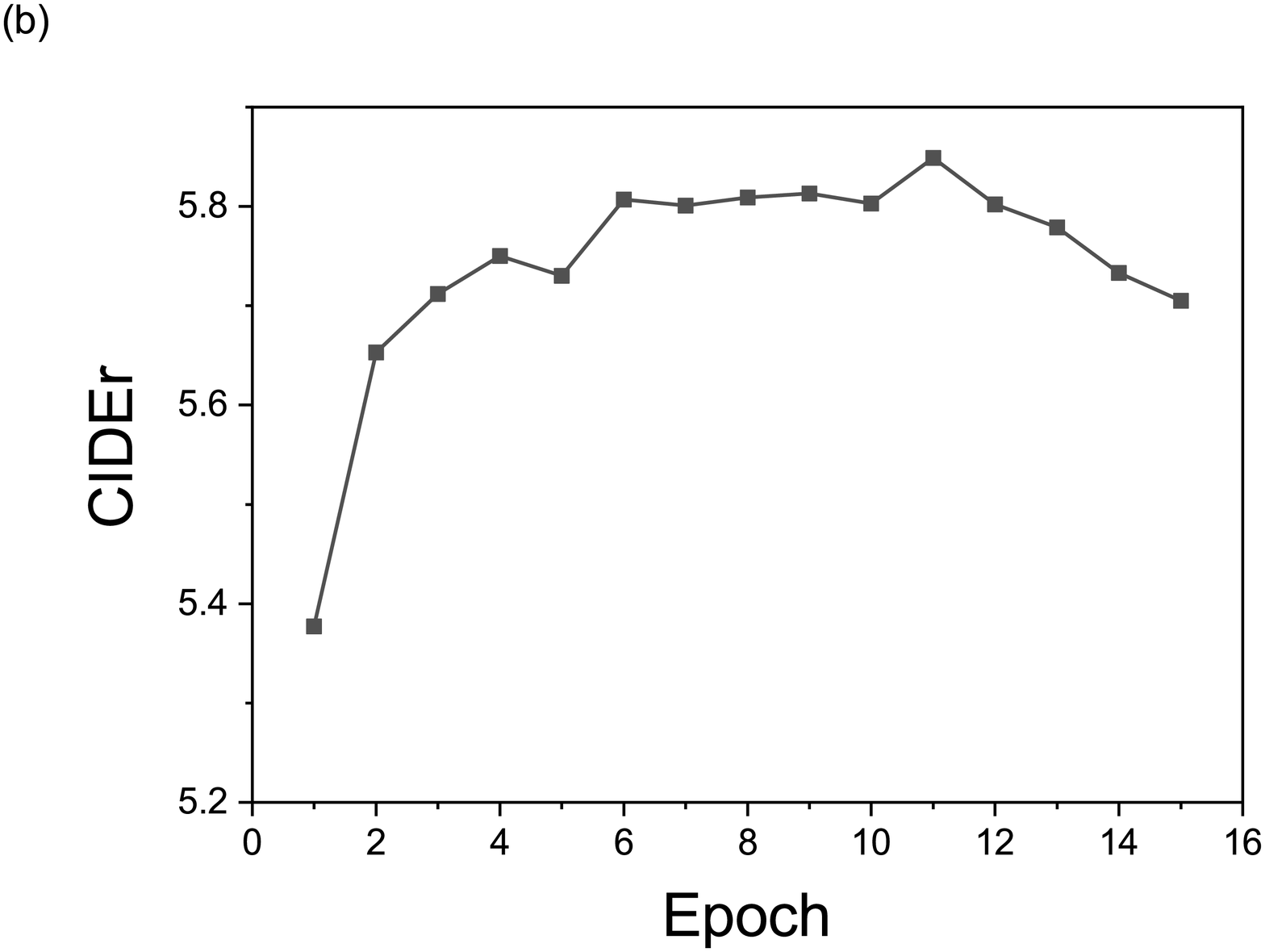}
	\caption{(a) The losses of the training and validation datasets as a function of epoch. (b) The CIDEr values of the testing dataset as a function of epoch.}
    \label{fig3}
    \end{center}
\end{figure}

Figure \ref{fig3}(a) shows the losses of the training and validation datasets as a function of the training epoch. It can be seen that the training loss keeps decreasing, while the validation loss saturates at about the 10{\it th} epoch, indicating that the generalization ability of the model reaches its maximum. Therefore, the parameters obtained at the 10{\it th} epoch are used by the model to generate the results in the following sections.

\begin{figure}[htbp]
	\begin{center}
	\includegraphics[width=0.78\textwidth]{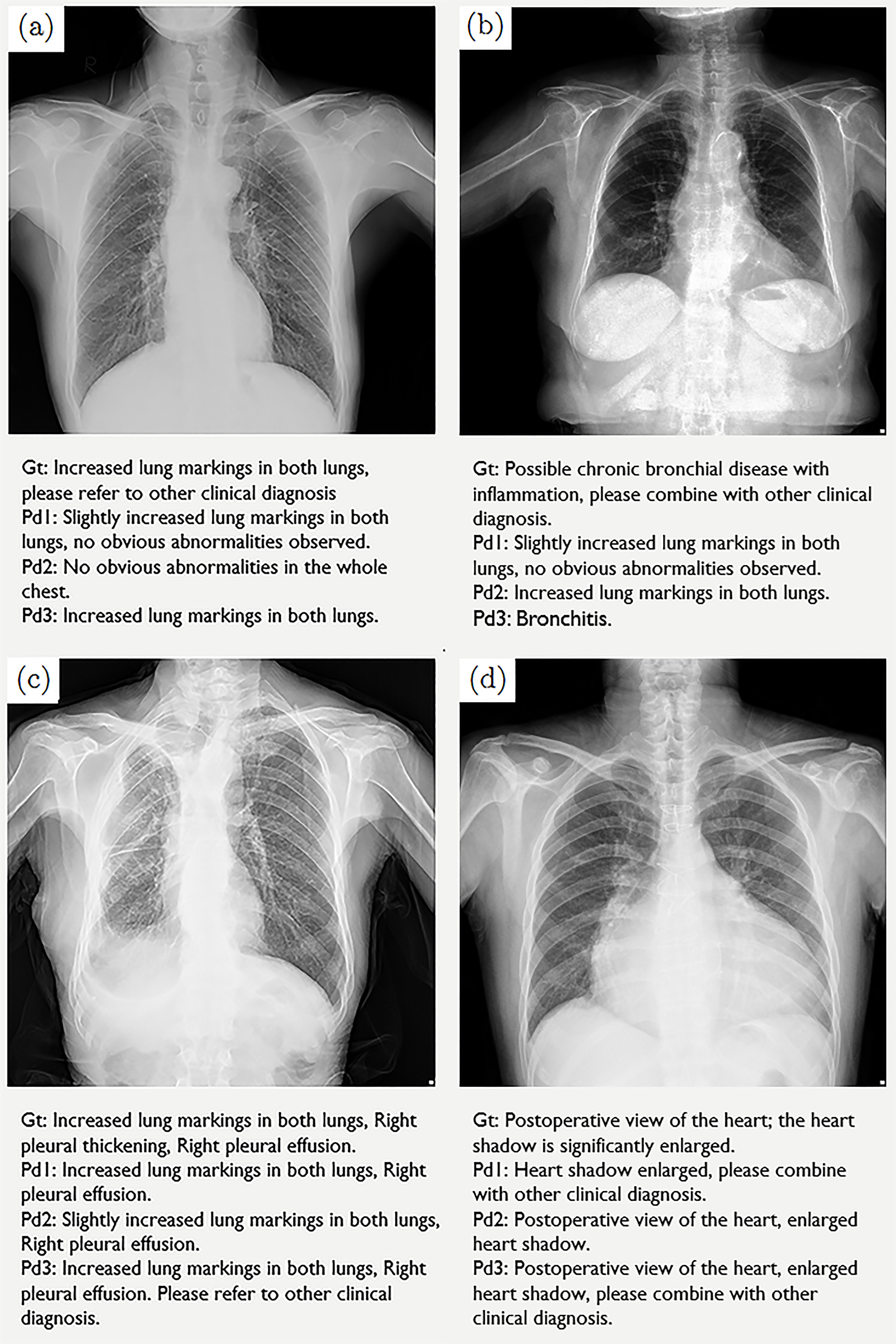}
	\caption{Several examples of the chest x-ray scans in the testing dataset and the corresponding descriptions generated automatically by the model. The descriptions are in Chinese and translated to English by human for the presentation purpose. The descriptions labeled by Gt are ground truth, and those labeled by Pd1, Pd2, and Pd3 are the top three descriptions based on their probabilities.
    \label{fig4}
}
	\end{center}
\end{figure}

Figure \ref{fig3}(b) shows the calculated CIDEr values for the testing dataset as a function of epoch. Note that for the testing set, the ground truth sentences were not used when generating the descriptions; they were only used to evaluate the descriptions after their generation. The Beam Search technique \cite{Sutskever2014Sequence} was used to generate multiple sentences for a given CXR image, and each sentence was assigned a preference probability. The top three sentences with the highest probabilities were recorded. Their CIDEr values were calculated against the ground truth and the highest one was used to calculate the curve shown in Fig. \ref{fig3}(b).
According to the figure, the average CIDEr value of the testing set increases as a function of the epoch, and saturates around 5.8 at the 10{\it th} epoch.
Fig. \ref{fig4}. shows several examples of the generated descriptions. For each scan, the top three predictions given by the model are shown, labeled as Pd1, Pd2, and Pd3, respectively, in the decreasing order of the preference probability. More examples are given in the supplemental materials. Fig. \ref{fig4}(a) shows a normal case with increased lung markings in both lungs. The model correctly recognizes the situation and generates descriptions with ``increased lung markings in both lungs'' in Pd1 and Pd3; while in Pd2, the model says ``no obvious abnormalities''.  Fig. \ref{fig4}(b) shows the scan of a patient with chronic bronchitis and inflammation, concluded based on the image as well as on the medical history. The model reports ``increased lung markings'' in Pd1 and Pd2, and directly gives ``bronchitis'' in Pd3, which is amazing since the model has no information of the medical history. Fig. \ref{fig4}(c) is a case with pleural effusion on the right side. The model identifies the symptom and correctly generates report for it. For the case in Fig. \ref{fig4}(d), the model correctly recognizes the increased size of the heart and also assumes the image to be a postoperative view. It is not known whether the model draws this conclusion based on the thin and bright strips near the heart region.

\begin{figure}[!ht]
\centering
\includegraphics[scale=0.6]{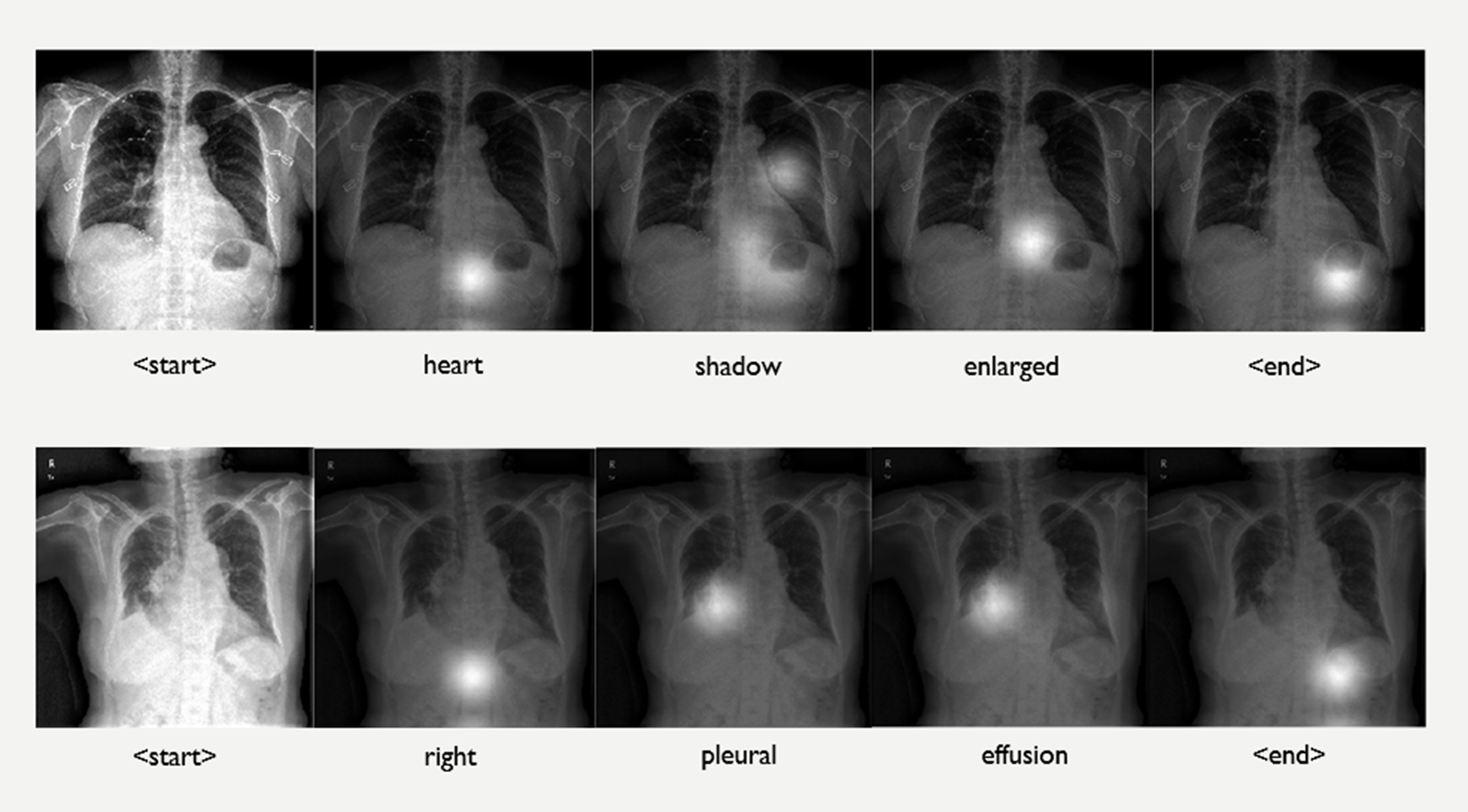}
\caption{Attention over time. As the network generates words step by step, its attention changes to different parts of the image, mimicking the visual behavior of human.}
\label{fig5}
\end{figure}

Figure \ref{fig5} shows the alignment of generated words with the relevant parts of the CXR images. In general, the alignments are consistent with human intuition. The alignments are enabled by the attention mechanism and provide another level of understanding of how the network works. They also facilitate debugging of the results.
The quality of the automatically generated reports were also evaluated by human experts. The procedure was as follows.
100 CXR scans were randomly extracted from the testing dataset and fed one by one to the neural network to generate reports. Another 100 CXR scans and the corresponding raw reports, which were written by human radiologists, were randomly extracted from the same dataset. These 200 scans and the associated reports were put together, shuffled and sent to experts for human inspection. Two radiologists (of associate chief physician level) were invited to examine the images and assess the quality of the associated reports, without knowing the origin of the reports - from human or machine. This was to prevent possible bias, either to human or to machine. The radiologists gave scores from 1 to 5 for each report, according to the standard as follows. An report with all conditions identified and accurately described was scored 5; an report with major conditions identified correctly but minor problems outside chest missed was scored 4, e.g., scoliosis, foreign matter in vitro; an report with major conditions identified correctly but minor problems inside the chest missed was scored 3, e.g., old lesions, fibrous stripes, post thoracic surgery, aortic calcification; if major conditions were identified but described inaccurately, a score 2 was given; If major conditions were missed or identified incorrectly, the score would be 1.

\begin{figure}[!ht]
\centering
\includegraphics[width=10cm,height=7cm]{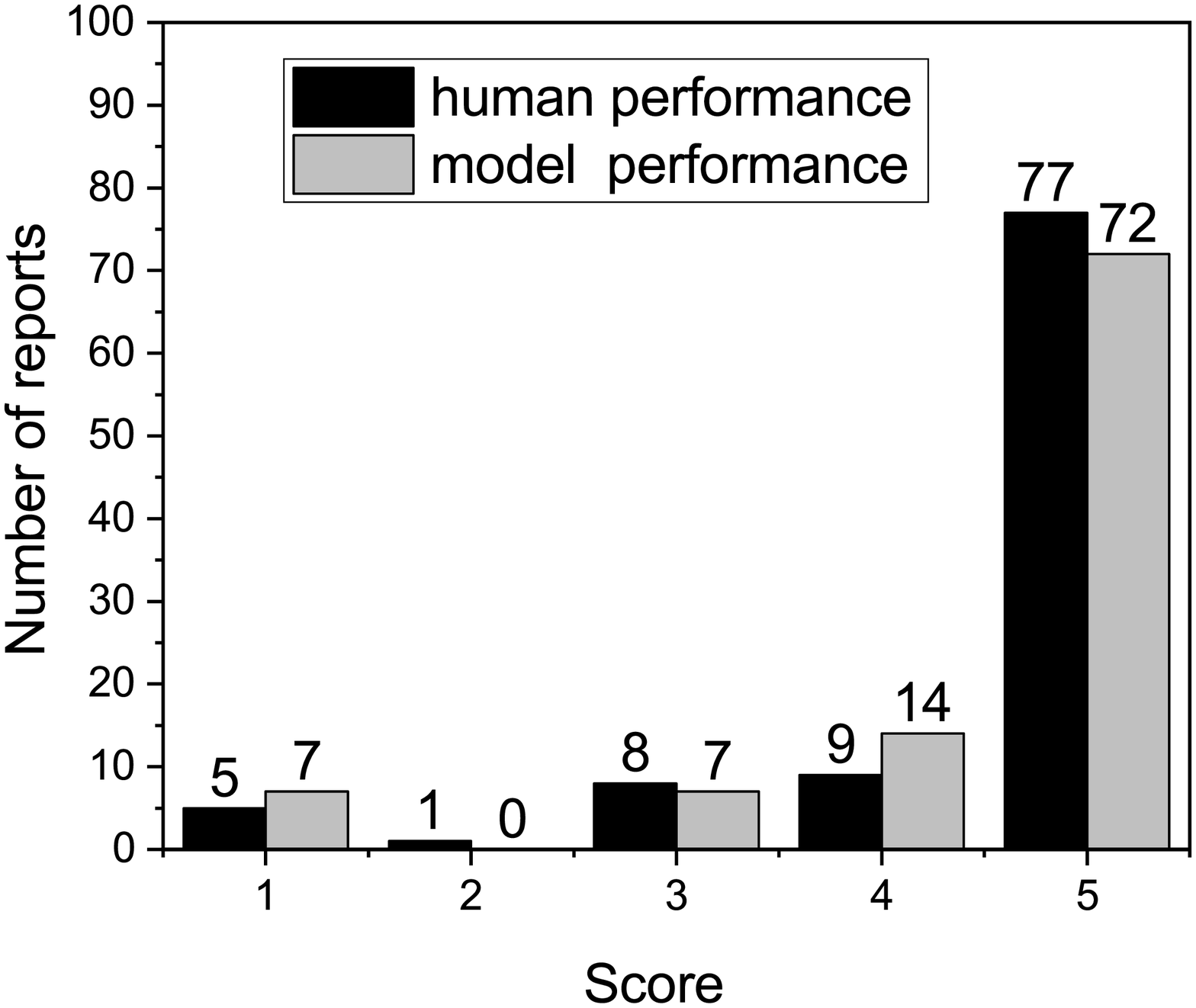}
\caption{Comparison of the quality of two groups of reports. In total there are 200 reports, half given by radiologists and half by the model. The quality of each report is measured by a score from 1 to 5, given by two radiologists independently without knowing the origin of the reports, from human or from the model.}
\label{fig6}
\end{figure}

Figure \ref{fig6} shows the score distributions for two groups of reports. It can be seen that for both groups, the majority are scored 5. For the group of reports given by human, 77\% are scored 5; while for the reports from the model, 72\% are scored 5. At the level of score 4, these two numbers are 9\% and 14\%, respectively. If we assume the scores of 4 or above are acceptable, then both groups have 86\% falling in this range, suggesting that the neural network is able to generate reports with quality comparable with that of human experts.

\section*{Discussion and Conclusion}

In summary, we developed a scheme that is able to automatically generate radiological reports/descriptions for given CXR, based on a deep convolutional neural network and an recurrent neural network with attention mechanism. We built a database containing more than 12,000 CXR scans and trained the model, and then evaluated the quality of the generated descriptions. The comparison between the generated descriptions and the ground truth  gave a CIDEr value of 5.8.
We also blended the generated descriptions with that given by radiologists, and invited other radiologists to score them. It was found that among the reports given by radiologists, 77\% received the highest score 5; while for the reports generated by the model, 72\% were scored 5. For the reports with score 4, the percentages were 9\% and 14\%, respectively. Therefore, the model is able to generate reports with high quality comparable to that of radiologists, and has the potential to be significantly improved as more training data are available.

The model developed here has some particular features. First, it learns from the raw radiological reports and is able to directly utilize the huge volume of CXR data generated in the hospital; no additional labeling work is required. This feature is particular useful since the acquisition of relevant annotations/labels is very expensive in the medical field.
Second, the model outputs description for a given CXR image instead of classifying it into a disease category. The consideration for this design is as follows. In clinical practice, it is not always feasible to draw solid conclusions on underlying diseases solely based on CXR images. For example, prominent/increased lung markings likely indicate an infection, chronic bronchitis, interstitial lung disease, heart failure, or normal aging. In case of such a symptom is observed, it is more appropriate to just describe the symptom, instead of giving a classification of diseases. The model is designed to follow this strategy. Moreover, this model behavior is similar to what radiologists usually do in their daily practice.

However, the model still requires improvement. Since the model is an end-to-end architecture that directly learns from reports and also generates reports, it does not explicitly give classification results. This makes it difficult to quantitatively evaluate the model performance. Currently we rely on human inspection for this purpose. We are dealing with this problem by adding a classification module in the neural network.

We believe the automated AI system developed in this work is useful and will greatly reduce the labor of doctors in the near future.

\section*{Ethics Committee Approvement}
The usage the above described chest X-ray data for this study has been approved by the Ethics Committee of Drum Tower Hospital, Nanjing University, under the document number 2019-100-01.

\section*{Role of the Funding Source}
The funding source has no influence on the scientific program and no role in the writing of this report or in the decision to submit it for publication.

\section*{Conflict of interests}
The authors declare no conflict of interests.

\section{References}
\bibliographystyle{elsarticle-num-names}

\newpage

\setcounter{figure}{0}
\renewcommand{\figurename}{Fig.S}
\begin{figure}[htb]
\centering
\includegraphics[width=1\textwidth]{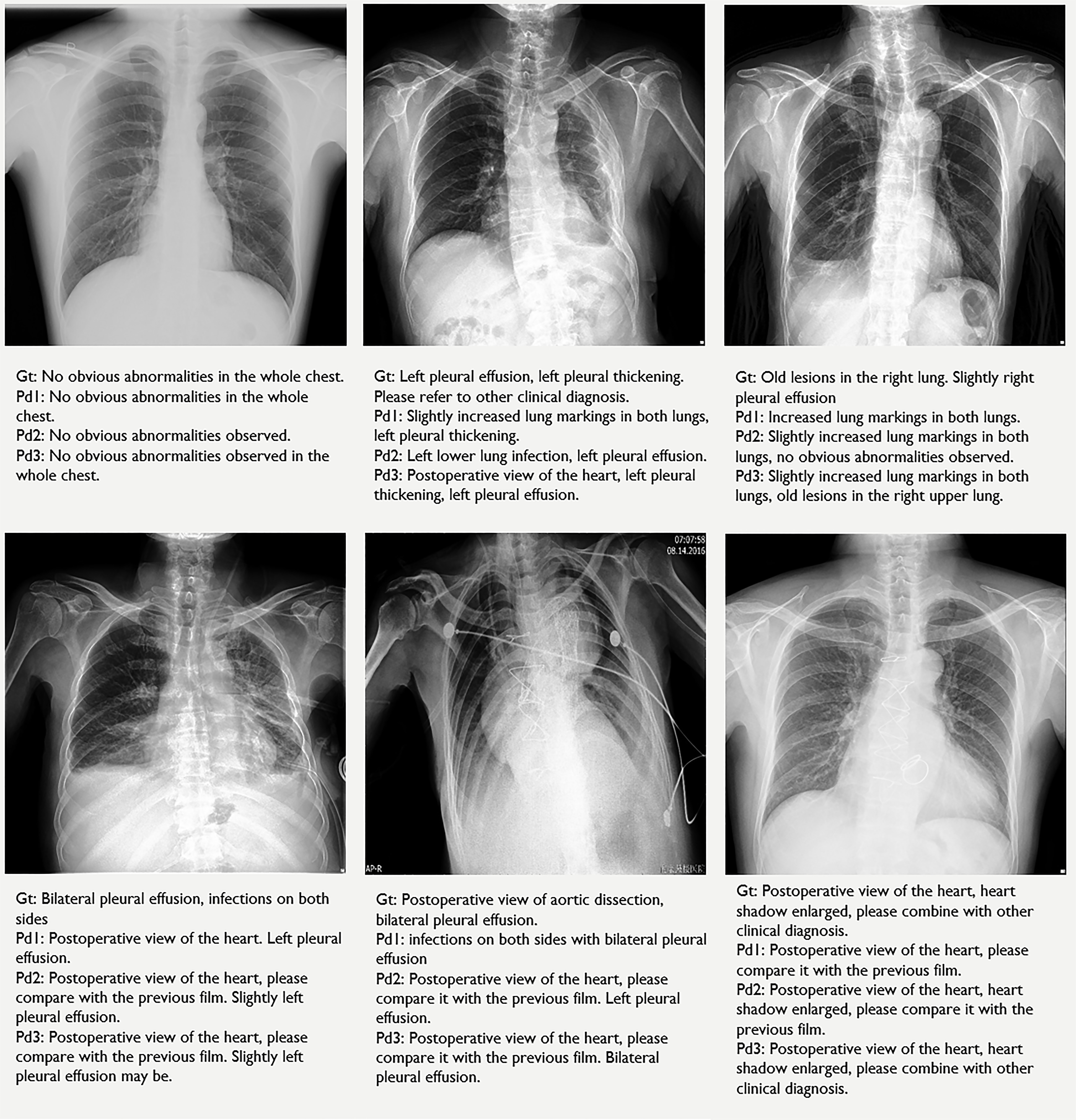}
\caption{More examples of the chest x-ray scans and the automatically generated descriptions in the testing set.}
\label{fig.S1}
\end{figure}

\end{document}